\documentclass[10pt,journal,compsoc]{IEEEtran}
\usepackage{epsfig}
\usepackage{graphicx}
\usepackage{siunitx}
\usepackage{balance}
\usepackage{color,soul} % this is added for \hl command
\usepackage{amsfonts} % added for \mathbb like Expectation
\usepackage{hyperref} % added for \href
\usepackage{multirow}
\usepackage{siunitx} % for comma-separated numbers 12,345
\usepackage{lscape}
\usepackage[skip=2pt]{caption}
\usepackage{booktabs}
\usepackage[flushleft]{threeparttable}

\ifCLASSOPTIONcompsoc
  \usepackage[nocompress]{cite}
\else
  \usepackage{cite}
\fi

\ifCLASSINFOpdf
\else
\fi

\begin{document}
\title{PrivacyNet: Semi-Adversarial Networks for Multi-attribute Face Privacy}
%
%

%\author{Vahid~Mirjalili, Sebastian~Raschka,Arun~Ross}

\author{Vahid Mirjalili\textsuperscript{1} \qquad Sebastian Raschka\textsuperscript{2}  \qquad Arun Ross\textsuperscript{1} \\
%\\ \;\;\;\;\;\; {\tt\small mirjalil@msu.edu} \;\;\;\; \; \; \; {\tt\small sraschka@wisc.edu}  \;\;\;  \;\;\;    {\tt\small rossarun@cse.msu.edu}  \\ \\
\textsuperscript{1}  Department of Computer Science and Engineering, Michigan State University\\
\textsuperscript{2}  Department of Statistics, University of Wisconsin -- Madison }

%,~\IEEEmembership{Member,~IEEE}% <-this % stops a space
%\IEEEcompsocitemizethanks{\IEEEcompsocthanksitem Corresponding author: Arun Ross\protect  ~~
%E-mail: rossarun@cse.msu.edu
%\IEEEcompsocthanksitem J. Doe and J. Doe are with Anonymous University.
%}% <-this % stops an unwanted space
%\thanks{Manuscript received January 19, 2018; revised August 26, 2015.}
%}

\markboth{} %IEEE Transactions on Image Processing,~Vol.~x, No.~x, August~201x}%
{Mirjalili \MakeLowercase{\textit{et al.}}: Semi-Adversarial Networks for Multi-attribute  Privacy}

\IEEEtitleabstractindextext{%
\begin{abstract}
Recent research has established the possibility of deducing soft-biometric attributes such as age, gender and race from an individual's face image with high accuracy. However, this raises privacy concerns, especially when face images collected for biometric recognition purposes are used for attribute analysis without the person's consent. To address this problem, we develop a technique for imparting soft biometric privacy to face images via an image perturbation methodology. The image perturbation is undertaken using a GAN-based Semi-Adversarial Network (SAN) - referred to as PrivacyNet - that modifies an input face image such that it can be used by a face matcher for matching purposes but cannot be reliably used by an attribute classifier. Further, PrivacyNet allows a person to choose specific attributes that have to be obfuscated in the input face images (e.g., age and race), while allowing for other types of attributes to be extracted (e.g., gender). Extensive experiments using multiple face matchers, multiple age/gender/race classifiers, and multiple face datasets demonstrate the generalizability of the proposed multi-attribute privacy enhancing method across multiple face and attribute classifiers. 
%In recent years, the utilization of biometric information has become more and more common for various forms of identity verification and user authentication. However, as a consequence of the widespread use and storage of biometric information, concerns regarding sensitive information leakage and the protection of users' privacy have been raised. Recent research efforts targeted these concerns by proposing the Semi-Adversarial Networks (SAN) framework for imparting gender privacy to face images. The objective of SAN is to perturb face image data such that it cannot be reliably used by a gender classifier but can still be used by a face matcher for matching purposes. In this work, we propose a novel Generative Adversarial Networks-based SAN model, PrivacyNet, that is capable of imparting {\em selective} soft biometric privacy to {\em multiple} soft-biometric attributes such as gender, age, and race. While PrivacyNet is capable of perturbing different sources of soft biometric information  reliably and simultaneously, it also allows users to choose to obfuscate specific attributes, while preserving others. The results from extensive experiments on five independent face image databases demonstrate the efficacy of our proposed model in imparting selective multi-attribute privacy to face images.

\end{abstract}

}

\maketitle

\IEEEdisplaynontitleabstractindextext

\IEEEpeerreviewmaketitle

\IEEEraisesectionheading{\section{Introduction}\label{sec:introduction}}

\IEEEPARstart{T}he use of automated methods to compare face images in order to determine the identity of an individual or to verify a claimed identity is known as face recognition~\cite{jain_introduction_2011}. Face recognition has been widely used in several applications, including, access control in smartphones, surveillance for public safety, and  finding missing children~\cite{wang_cosface_2018,trigueros_face_2018,foresti_face_2003,kamgar_toward_2011,grgic_scface_2011,qezavati_partially_2019,kwon_biometric_2008,del_automated_2016,bobak_solving_2016,zhao_dynamic_2007}. Examples of face recognition methods include Elastic Bunch Graph Matching~\cite{wiskott_face_1997}, Active Appearance Models~\cite{hong_face_2008}, Sparse Representation~\cite{wright_robust_2008}, as well as more recent techniques based on Deep Learning~\cite{taigman2014deepface}.
%~\cite{xi_local_2016,masi_deep_2018}.

The primary purpose of collecting and storing face images in a biometric system is for the recognition of individuals. Yet, face images stored in a database implicitly contain auxiliary information about each individual~\cite{dantcheva_what_2016,GonzalezTIFS2018}. These auxiliary information, sometimes referred to as \emph{soft biometric attributes}, include gender, age, ethnicity, body mass index, and health characteristics~\cite{dantcheva_what_2016,NixonSoftBiometrics_2015,SunDemographicPAMI_2018}.  

Soft biometrics can facilitate a large variety of applications, such as improving face recognition performance, clustering users, or developing targeted advertisements~\cite{ReidSoftBiometricsPAMI_2014,martinho2016categorising,jaha_augmenting_2019}. Recent advances in machine learning has made it possible to extract such soft biometric attributes from face images automatically and, in many cases, with a high degree of accuracy~\cite{dantcheva_bag_2011,NixonSoftBiometrics_2015,dantcheva_what_2016}. However, users of such biometric systems may prefer not to be profiled based on their demographic attributes and may wish to opt-out of such services due to privacy concerns~\cite{cynthia_privacy_2018,cynthia_differential_2019,cynthia_individual_2020}. In this regard, certain privacy laws allow users to choose what information about themselves to reveal and what information to conceal~\cite{kindt_privacy_2016,wu_privacy_2019,sun_hybrid_2018,sadeghi_imparting_2020}. Moreover, in the near future, biometric applications are expected to implement best-practices with regard to respecting the privacy of users by preventing automatic information extraction from face images in the absence of users' consent~\cite{medcn_selective_2018,yang_using_2018,cynthia_privacy_2018}. However, even if information is not extracted intentionally, user images stored in a database are still susceptible to privacy breaches via third party users or applications~\cite{jasserandPhDThesis_2019}. Thus, to provide actionable means and guarantees for preventing the automatic mining of personal information from face images, recent research has explored the possibility of imparting soft biometric privacy to face images by modifying the image data directly~\cite{othman_privacy_2014,sim_controllable_2015,mirjalili_semi_2018,chhabra_anonymizing_2018,suo_high-resolution_2011_long,tripathy_privacy_2019}.

%For example, Google's privacy policy includes a statement indicating that the personal information being collected and utilized is willingly provided by the users, and no personal information is being extracted from other sources using data mining techniques~\cite{google_privacy_2019}.

To provide a practical approach for imparting gender privacy to face images, Mirjalili et al. previously developed the Semi-Adversarial Network (SAN) model~\cite{mirjalili_semi_2018}.  This SAN model is able to conceal gender information in face images while retaining satisfactory face matching accuracy. In later studies, improvements of the SAN model resulted in state-of-the-art matching performance with arbitrary face matchers under the constraint that arbitrary gender classifiers were not able to extract gender information from the modified face images~\cite{mirjalili_flowsan_2019,mirjalili_gender_2018}. 

\begin{figure}
\begin{center}
   \includegraphics[width=1\linewidth]{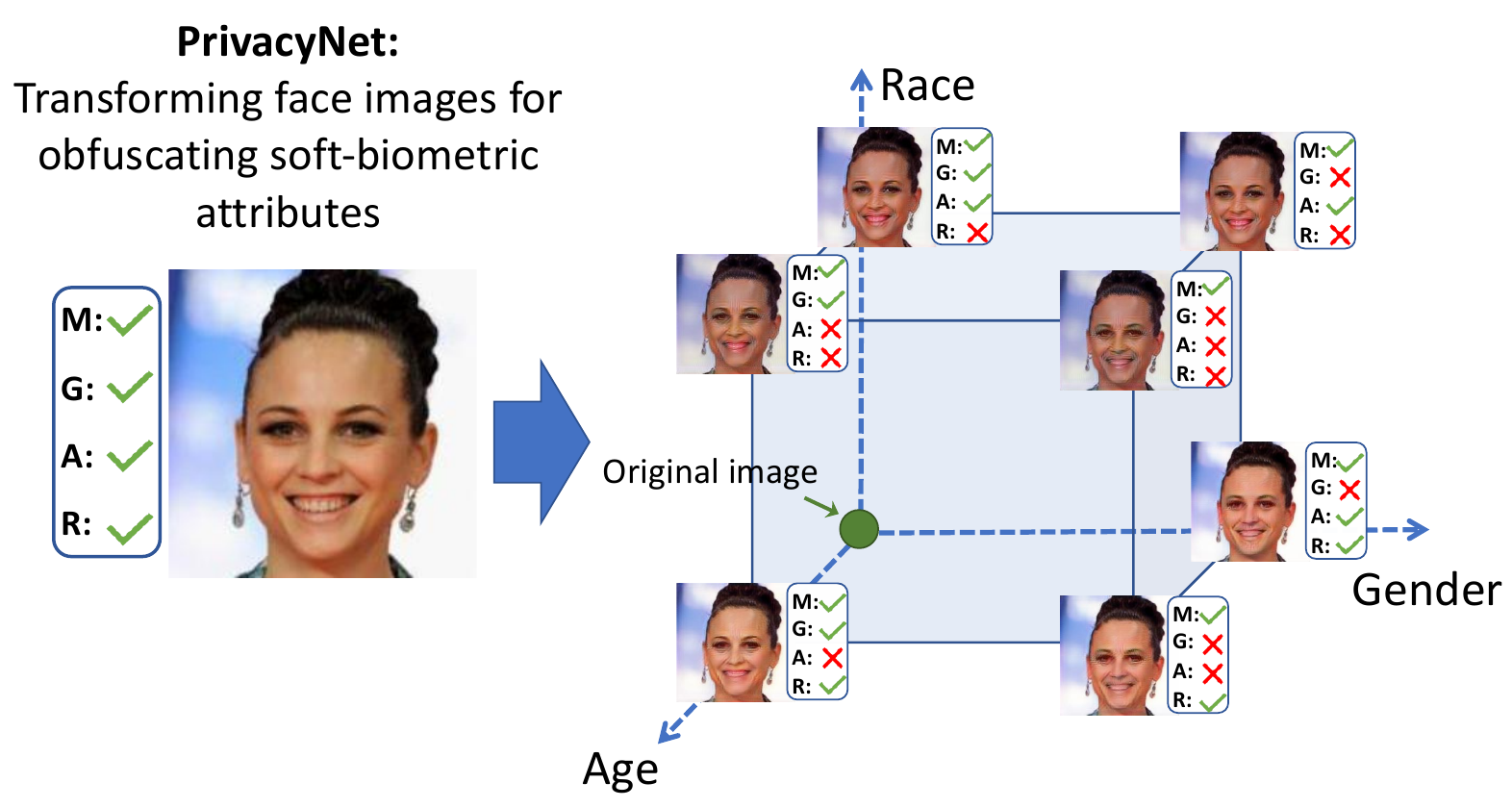}
\end{center}
   \caption{Illustration of the overall objective of this work: transforming an input face image across three orthogonal axes for imparting multi-attribute privacy selectively while retaining biometric recognition utility. The abbreviated letters are M: Matching, G: Gender, A: Age, and R: Race. }
   %\hl{*This fig needs to be referenced in the text; maybe later in the intro.}}
\label{fig:overall-idea}
\end{figure}

While the previously developed SAN model is only capable of hiding gender information, in this paper, we propose a new model, named {\em PrivacyNet},  for imparting multi-attribute privacy to face images that includes gender,\footnote{The terms `gender' and `sex' have been used interchangeably in the biometric literature. It must be noted that gender is a social or cultural construct, while sex is based on biological characteristics.} race,\footnote{The terms `race' and `ethnicity' have been used interchangeably in the biometric literature. An exact definition of either of these two terms appears to be debatable.} and age. The overall objective of this work is to develop a model that can induce selective (which attributes to conceal) and collective (how many attributes to conceal) perturbations to impart soft biometric privacy to face images (Fig.~\ref{fig:overall-idea}) while retaining biometric matching performance. The main differences between PrivacyNet and previous work on perturbing facial attributes while maintaining biometric matching utility are as follows:
\begin{itemize}
    \item The adversarial examples generated in~\cite{mirjalili_soft_2017} were only focused on the gender attribute, and the resulting outputs were not generalizable to unseen\footnote{The term ``unseen" refers to matchers, classifiers or data that were not used during training.} gender classifiers.
    \item The generalizability issue of~\cite{mirjalili_soft_2017} was addressed by developing semi-adversarial networks~\cite{mirjalili_semi_2018} and their subsequent variants ~\cite{mirjalili_gender_2018} and ~\cite{mirjalili_flowsan_2019}. However, all of the aforementioned methods are focused on a single attribute, gender, whereas PrivacyNet can perturb race, gender, and age attributes.
    \item Furthermore, PrivacyNet is a GAN-based model with cycle consistency loss, in contrast to the regular convolutional autoencoders used in~\cite{mirjalili_semi_2018,mirjalili_gender_2018,mirjalili_flowsan_2019}
\end{itemize}

The organization of this paper is as follows. Section~\ref{sec:related-work} discusses existing methods for imparting privacy to face images. In Section~\ref{sec:proposed-method}, we describe our proposed PrivacyNet method, including a detailed description of the neural network architecture and datasets used in this study. Section ~\ref{sec:exp-results} presents and discusses the experimental results obtained by analyzing the matching utility of face images after perturbing facial attributes.
%The organization of this paper is as follows. In Section~\ref{sec:related-work} the existing methods for imparting privacy to face images are provided. Then, our proposed work is described in Section~\ref{sec:proposed-method}. Finally, we conclude with the experimental results and their discussion in Section~\ref{sec:exp-results}.}

\section{Related Work}\label{sec:related-work}

With recent advances in machine learning and deep learning for computer vision, the prediction of soft biometric attributes such as age, gender, and ethnicity from facial biometric data has been widely studied~\cite{niu_ordinal_2016,cao_coral_2019,gunther_affact_2017,dantcheva_what_2016,bobeldyk_predicting_2019}. For instance, the use of convolutional neural networks for predicting the gender from face images has resulted in models with almost perfect prediction accuracy~\cite{levi_age_2015,mansanet_local_2016,jia_gender_2016,castrillon_descriptors_2017,gunther_affact_2017}. Methods for estimating the apparent age from face images are similarly well studied, and current-state of the art methods can predict the apparent age of a person with a prediction error below three years on average~\cite{niu_ordinal_2016,chen_cascaded_2016,chen_using_2017,cao_coral_2019}. 

While tremendous progress has been made towards the automatic extraction of personal attributes of face images, the development of methods and techniques for imparting soft biometric privacy is still a relatively recent area of research. 
In 2014, Othman and Ross introduced the concept of soft biometric privacy, where a face image is modified such that the gender information is confounded while the recognition utility of the face image is preserved~\cite{othman_privacy_2014}. The researchers proposed a face mixing approach, where a face image is morphed with a candidate face image from the opposite gender. As a result, the resulting mixed face image contains both male and females features such that the gender information was fully anonymized. 
Sim and Zhang then developed methods for imparting soft biometric privacy to multiple attributes based on multi-modal discriminant analysis, in which certain attributes can be selectively suppressed while retaining others~\cite{sim_controllable_2015}. They proposed a technique that decomposes a face image representation into orthogonal axes corresponding to gender, age, and ethnicity, and the identity information is left as a residual of this decomposition. This enables transforming a face image along one axis resulting in modifying the corresponding attribute, while other information of the face image remains visibly unchanged to the human eye. They also showed that their proposed method can alter identities of face images, which is useful for face de-identification~\cite{gross_model-based_2006_long}. However, Sim and Zhang's~\cite{sim_controllable_2015} method cannot explicitly preserve the matching performance of transformed face images and, therefore, the biometric utility of the resulting face images is severely diminished.

In 2013, Szegedy et al.~\cite{szegedy_intriguing_2013} studied the vulnerability of Deep Neural Networks (DNNs) towards adversarial perturbations. Adversarial perturbations are small perturbations added to an input image, typically imperceptible to a human observer, that can cause the DNN to misclassify images with high confidence.  In recent years, several methods for generating such adversarial perturbations have been proposed, and the development of methods that make DNN-based models more robust against these so-called adversarial attacks remains an active area of research~\cite{oleszkiewicz_siamese_2018,akhtar_threat_2018,goodfellow_explaining_2014,poursaeed_generative_2018,moosavi_deepfool_2016,su_cleaning_2019,taheri_razornet_2019,athalye_synthesizing_2018}.
The vulnerability to adversarial attacks raises several security concerns for the use of machine learning systems in computer vision applications~\cite{graese_assessing_2016,yang_using_2018,oleszkiewicz_siamese_2018,ross_some_2019}. Recently, Rozsa et al.~\cite{rozsa_facial_2016} investigated the robustness of binary facial attribute classifiers to adversarial attacks. 
Based on the concept of adding adversarial perturbations to an input image, Mirjalili and Ross~\cite{mirjalili_soft_2017} investigated the possibility of generating adversarial perturbations for imparting soft-biometric privacy to face images. This scheme was further extended by Chhabera et al.~\cite{chhabra_anonymizing_2018} to conceal multiple face attributes simultaneously. 
While these perturbation-based methods are shown to successfully derive adversarial examples based on a specific attribute classifier, the perturbed output images are not generalizable across unseen attribute classifiers. For a real-world privacy application, generalizability of adversarial examples to unseen attribute classifiers is critical~\cite{mirjalili_gender_2018}.

Recently, methods have been developed that impart privacy through the design and use of specific face representation vectors, which have been derived from the original face images without including the sensitive information that is to be concealed~\cite{xie_controllable_2017,terhorst_unsupervised_2019,morales_sensitivenets_2019,roy_mitigating_2019,sadeghi_global_2019}. For instance, the  SensitiveNet~\cite{morales_sensitivenets_2019} model generates agnostic face representations for biometric recognition such that gender and race information are removed from these representations~\cite{jia_right_2018}. However, storing face representation vectors may not be desirable in many applications since these vectors are neither interpretable by humans nor compatible with future face recognition software. In this work, we develop a generally applicable method that applies perturbations to the face images directly instead of the derived representations. 

In previous work~\cite{mirjalili_semi_2018}, Mirjalili et al. developed a deep learning-based model to generate perturbed examples for obfuscating gender information in face images. The neural network was coined Semi-Adversarial Network (SAN) and is composed of a convolutional autoencoder for synthesizing face images such that the gender information in the synthesized images is obfuscated while their matching utility is preserved. The SAN model is trained using an auxiliary gender classifier and an auxiliary face matcher. After training, the auxiliary subnetworks are discarded and the convolutional autoencoder is used for performance evaluation on unseen data. It was shown that this model is able to suppress gender information as assessed by some {\em unseen}\footnote{In contrary to ``auxiliary'' classifiers, the  term  ``unseen''  indicates  that  the  attribute classifier  (or  face  matcher) was not used during the training stage.} attribute classifiers while the matching utility, assessed by unseen face matchers, was retained. Moreover, the generalizability of SAN models to fool arbitrary gender classifiers can be further enhanced by diversifying the auxiliary classifiers during training~\cite{mirjalili_gender_2018} or by combining multiple, diverse SAN models~\cite{mirjalili_flowsan_2019}.

%While previous work~\cite{mirjalili_flowsan_2019} has shown its efficacy in imparting gender privacy to face images, existing methods for imparting multi-attribute privacy are  limited. As  mentioned earlier, the work of Chhabra et al.~\cite{chhabra_anonymizing_2018} is {\em not generalizable} to previously unseen attribute classifiers. Furthermore, the controllable face privacy model proposed by Sim and Zhang~\cite{sim_controllable_2015} cannot retain the recognition utility of the face images. 

%Apart from imparting soft biometric privacy via adversarial-based SAN-based models, and its variants have shown remarkable performance in many computer vision tasks such as image-to-image translation~\cite{isola_image2image_2017} and face image synthesis~\cite{choi_stargan_2017,zhu_unpaired_2017,kim_learning_2017,antipov_face_2017,shocher_semantic_2020,jiang_psgan_2020}. 

With the development of Generative Adversarial Networks (GANs)~\cite{goodfellow_generative_2014_long,lu_recent_2017,raschka_python_2019}, different GAN-based models for editing facial attributes have been proposed in the literature~\cite{he_attgan_2019,zhang_generative_2018,song_face_2020,shen_learning_2017,zhou_genegan_2017}. These methods focus on selected face attributes such as facial hair, eyeglasses, hair color and skin-tone, which can be modified without significantly affecting other attributes and biometric recognition capabilities. The aforementioned facial attributes are only loosely connected to the identity information in the face image. On the other hand, gender and race are intricately tied to the identity of the subjects, and, therefore, cannot be easily modified without affecting the recognition capabilities of face matchers. Consequently, the aforementioned GAN models and the PrivacyNet method we propose in this paper have different objectives. To the best of our knowledge, the only GAN-based model that has a similar objective as PrivacyNet is the Deep Identity-Aware Transfer (DIAT) model ~\cite{li_deep_2016}. Besides transferring attributes that are loosely connected with identity, DIAT also considers gender transfer. DIAT relies on a perceptual loss via a pre-trained VGG-face~\cite{cao_vggface2_2018} network to preserve the recognition performance of the attribute-transferred output images. However, their experimental results still showed significant decrease in face verification performance after transferring the eyeglass attributes of the face images. %GAN models are not considered as a viable solution for imparting soft biometric privacy since the objective of GAN-based models for image-to-image translation is to synthesize realistic face images while the biometric matching utility is not explicitly preserved.}

%While our previous work~\cite{mirjalili_flowsan_2019} has shown its efficacy in imparting gender privacy to face images, existing methods for imparting multi-attribute privacy are  limited. As  mentioned earlier, the work of Chhabra et al.~\cite{chhabra_anonymizing_2018} is {\em not generalizable} to previously unseen attribute classifiers. Furthermore, the controllable face privacy model proposed by Sim and Zhang~\cite{sim_controllable_2015} cannot retain the recognition utility of the face images. 

The main contribution of this work is the design of a {\bf multi-attribute} face privacy model to provide {\bf controllable soft-biometric privacy}. This proposed GAN-based privacy model, which we refer to as ``PrivacyNet,''  modifies an input face image to {\bf obfuscate soft-biometric attributes} while {\bf maintaining the recognition capability} on the generated face images. To the best of our knowledge, the ``PrivacyNet'' model proposed in this paper is the first method for  multi-attribute privacy that generalizes to unseen attribute classifiers while preserving the recognition utility of face images across unseen face matchers.

\section{Proposed method}\label{sec:proposed-method}

\begin{figure*}%[t!]
\begin{center}
   \includegraphics[width=0.99\linewidth]{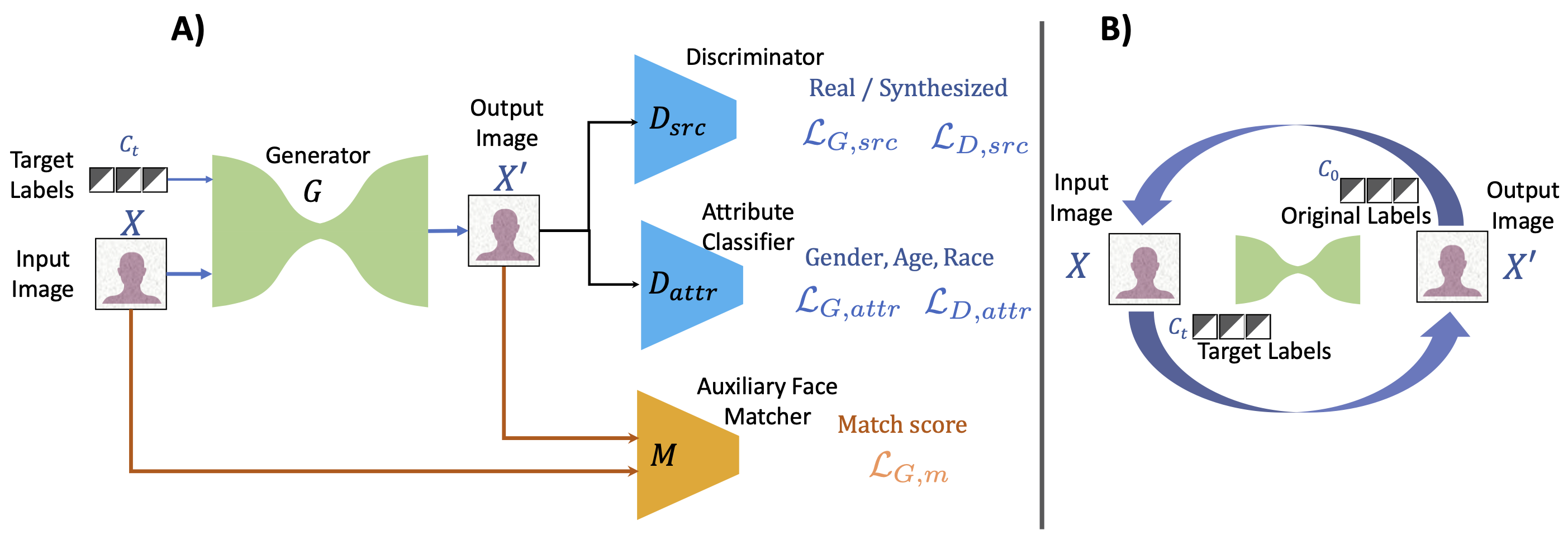} %general-architecture.png}
\end{center}
   \caption{Schematic representation of the architecture of PrivacyNet for deriving perturbations to obfuscate three attribute classifiers -- gender, age and race -- while allowing biometric face matchers to perform well. (A) Different components of the PrivacyNet: generator, source discriminator, attribute classifier, and  auxiliary face matcher. (B) Cycle-consistency constraint applied to the generator by transforming an input face image to a target label and reconstructing the original version.}
\label{fig:general-arch}
%\label{fig:onecol}
% \{\hl{Vahid will add the math. abbrev. such as $D_{attr}$, $\mathcal{R}_M$ etc. to this figure.}\}
\end{figure*}

\subsection{Problem Formulation}\label{sec:problem-form}

%Semi-Adversarial Network (SAN)~\cite{mirjalili_semi_2018} was first developed as for multiple attributes as described in the next section.
Given a face image $X$, let  $S_{\text{obf}}$ be a set of face attributes to be obfuscated and $S_{\text{keep}}$ be a set of attributes to be preserved. The overall objective is to find function $\phi$ that applies some perturbations to the input image $X$ such that  $X^\prime=\phi(X)$ has the following properties:

\begin{itemize}
\item For a soft biometric attribute $a\in S_{\text{obf}}$, the performance of an unseen attribute classifier $f_a$ is substantially reduced. 

\item For the remaining set of attributes $b\in S_{\text{keep}}$, the performance of an arbitrary classifier $f_b$ is not noticeably adversely affected; that is, the performance of an attribute classifier $f_b$ on perturbed image $X^\prime$ is close to its performance on the original face image $X$.

\item The primary biometric utility, which is face recognition, must be retained for the modified face image, $X^\prime$. In other words, given pairs of image examples before ($\langle X_1,X_2 \rangle$) and after ($\langle X_1^\prime,X_2^\prime \rangle$) perturbations, the matching performance as assessed by an arbitrary face matcher ($f_M$) is not substantially affected, i.e., \\ $f_M(X_1,X_2) \approx f_M(X_1^\prime, X_2^\prime) \approx f_M(X_1,X_2^\prime) \approx f_M(X_1^\prime,X_2)$.
\end{itemize}

\subsection{PrivacyNet}

\begin{figure*}
\begin{center}
   \includegraphics[width=0.99\linewidth]{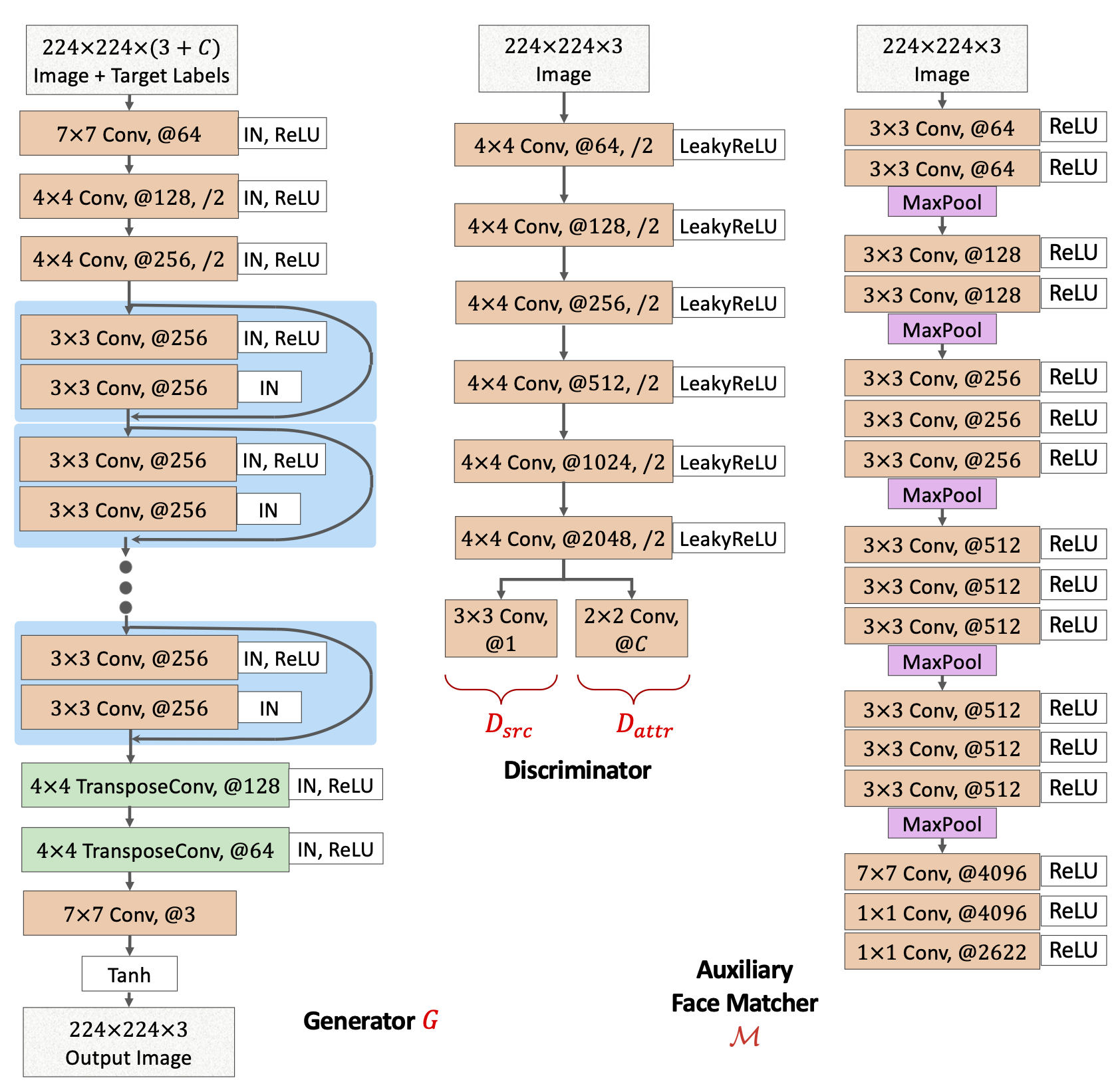} %general-architecture.png}
\end{center}
   \caption{The detailed neural network architecture of the four sub-networks of PrivacyNet: the generator $G$, the discriminators $D_{src}$ and $D_{attr}$, and the pre-trained auxiliary face matcher ${\mathcal{M}}$. Note that $D_{src}$ and $D_{attr}$ share the same convolutional layers and only differ in their respective output layers.}
\label{fig:privacynet-arch}
%\label{fig:onecol}
\end{figure*}

According to the objectives described in Section~\ref{sec:problem-form}, the PrivacyNet neural network architecture (Fig.~\ref{fig:general-arch}A) is composed of four sub-networks: A generator ($G$) that modifies the input image, a source discriminator ($D_{src}$) which determines if an image is real or modified, an attribute classifier ($D_{attr}$) for predicting facial attributes, and an auxiliary face matcher ($M$) for biometric face recognition. Along with the input image, both generator and discriminator receive the attribute labels as conditional variables, that are spanned to the same width and height as the input image, ($224\times 224$). Together, these subnetworks form a cycle-consistent GAN~\cite{zhu_unpaired_2017} as illustrated in Fig.~\ref{fig:general-arch}B. 
Given an RGB input face image $X$, the attribute label vector $\mathcal{V}_0 \in \mathbb{Z}^c$ corresponds to the ground truth attribute labels of the original face image. The target label vector $\mathcal{V}_t \in \mathbb{Z}^c$ ($c$ is the total number of attributes) denotes the desired facial attributes for modifying the face image. Given a target vector $\mathcal{V}_t \neq \mathcal{V}_0$, the objective of the generator $G$ is to synthesize a new image $X^\prime = G(X, \mathcal{V}_t)$ such that $X^\prime$ is mapped to the target label vector $\mathcal{V}_t$ by an attribute classifier $D_{attr}$. 
The other component of the GAN model is a source discriminator $D_{src}$, which is trained to distinguish real images from those synthesized by the generator.

The total loss terms for training the discriminator ($\mathcal{L}_{D,tot}$) and the generator ($\mathcal{L}_{G,tot}$) are as follows:

\begin{equation}
\mathcal{L}_{D,tot} = \mathcal{L}_{D,src} + \lambda_{D,attr} \mathcal{L}_{D,attr},
\end{equation}

\noindent and

\begin{equation}
\begin{array}{rl}
\mathcal{L}_{G,tot} = & \mathcal{L}_{G,src} + \lambda_{G,attr} \mathcal{L}_{G,attr} + \\
&\lambda_m \mathcal{L}_{G,m} + \lambda_{rec} \mathcal{L}_{G,rec},
\end{array}
\end{equation}

\noindent where, $\lambda$ coefficients are hyperparameters representing the relative weights for the corresponding loss terms. The individual terms of the total loss for the discriminator ($\mathcal{L}_{D,tot}$) and the generator ($\mathcal{L}_{G,tot}$) are described in the following paragraphs.

 For the discriminator, the loss term associated with source discrimination (i.e., discriminating between real and synthesized images) is defined as
\begin{equation}\label{eq:loss_d_src}
\begin{array}{rl}
    \mathcal{L}_{D,src} = & \mathbb{E}_{X,\mathcal{V}_0}\Big[-\log \big(D_{src}(X,\mathcal{V}_0)\big)\Big] +\\
    & \mathbb{E}_{X, \mathcal{V}_t}\Big[-\log\big(1-D_{src}\left(G(X, \mathcal{V}_t),\mathcal{V}_t\right)\big)\Big],
 \end{array}
\end{equation}

\noindent where, $\mathbb{E}_{X,\mathcal{V}}\big[f(X,\mathcal{V})\big]$ represents the expected value of the random variable $f(X,\mathcal{V})$ taken over distribution of $X$ given the conditional variable $\mathcal{V}$. Similarly, the loss associated with the source discrimination for the generator subnetwork is defined as 

 \begin{equation}\label{eq:loss_g_src}
    \mathcal{L}_{G,src} = \mathbb{E}_{X, \mathcal{V}_t}\Big[\log(1-D_{src}\big(G(X, \mathcal{V}_t),\mathcal{V}_t)\big)\Big],
\end{equation}

\noindent where, $D_{src}(X)$ returns the estimate of the probability that the input image $X$ is real or was synthesized by the generator. 

Next, the loss terms for attribute classification are defined as 

\begin{equation}\label{eq:loss_d_attr}
    \mathcal{L}_{D,attr} = \displaystyle \mathbb{E}_{X, \mathcal{V}_0}\Big[-\log\big( D_{attr}(\mathcal{V}_0 \vert X)\big)\Big]
\end{equation}

and

\begin{equation}\label{eq:loss_g_attr}
    \mathcal{L}_{G,attr} = \displaystyle \mathbb{E}_{X, \mathcal{V}_t}\Big[-\log\big( D_{attr}(\mathcal{V}_t \vert G(X, \mathcal{V}_t))\big)\Big],
\end{equation}

\noindent where, $D_{attr}(\mathcal{V}|X)$ is the probability that input image $X$ belongs to attribute class $\mathcal{V}$. 

The loss term for optimizing the performance of the biometric face matcher $\mathcal{M}$ on the perturbed images is defined as the squared $L_2$ distance between the normalized features of the original face image $X$ and those of the synthesized image $G(X, \mathcal{V}_t)$: 

\begin{equation}\label{eq:loss_g_rec}
    \mathcal{L}_{G,m} = \mathbb{E}_{X, \mathcal{V}_t}\Big[\|R_\mathcal{M}(X) - R_\mathcal{M}(G(X, \mathcal{V}_t))\|_2^2\Big],
\end{equation}

\noindent where, $R_{\mathcal{M}}(X)$ is the normalized face descriptor of face image $X$ after applying a face matcher $\mathcal{M}$.

Lastly, a reconstruction loss term is used to form a cycle-consistent GAN that is able to reconstruct the original face image $X$ from its modified face image $X'=G(X, \mathcal{V}_t)$:
\begin{equation}\label{eq:l1-reconstrution}
    \mathcal{L}_{G,rec} = \mathbb{E}_{X,\mathcal{V}_0, \mathcal{V}_t} \Big[\|X - G(G(X, \mathcal{V}_t), \mathcal{V}_0)\|_1\Big].
\end{equation}

Note that the distance term in Eq.~\ref{eq:l1-reconstrution} is computed as the pixel-wise $L_1$ norm between the original and modified images, which empirically results in less blurry images compared to employing a $L_2$ norm as the distance measure~\cite{isola_image2image_2017}.

\subsection{Neural Network Architecture of PrivacyNet}

The composition of the different neural networks used in PrivacyNet, generator $G$, real vs. synthetic classifier $D_{src}$, attribute classifier $D_{attr}$, and face matcher $R_{\mathcal{M}}$ is described in 
Fig.~\ref{fig:privacynet-arch}. The generator and the discriminator architectures were adapted  from~\cite{choi_stargan_2017} and~\cite{zhu_unpaired_2017}, respectively.

{\bf Generator.} The generator $G$ receives as input an RGB face image $X$ of size $224\times 224\times 3$ along with the target labels $\mathcal{V}_t$ concatenated as extra channels. The first two convolutional layers, with  stride $2$, reduce the size of the input image  to a to $32\times 32$ with 128 channels. The convolutional layers are followed by instance normalization layers (InstanceNorm)~\cite{ulyanov_instance_2016}. The layer activations are computed by applying the non-linear ReLU activation function to the InstanceNorm outputs. Then, $6$ residual blocks~\cite{he_deep_2016} are applied, followed by two transposed convolution for upsampling the image size to $224\times 224$. Finally, the output image $X'$ is constructed by a $1\times 1$ convolution layer and the hyperbolic tangent ($Tanh$) activation function, which returns pixels in the range $(-1, 1)$ (the input image pixels are also scaled to be in range $[-1, 1]$).% the scaling of the input images is described in Sec.~\ref{sed:datasets}). 

{\bf Discriminator and Attribute Classifier.} The discriminator, as shown in Fig.~\ref{fig:privacynet-arch}, combines the source discriminator $D_{src}$ and the attribute classifier $D_{attr}$ into one network where all the layers except the last convolution layer are shared among the two tasks. All the shared convolution layers are followed by a Leaky ReLU non-linear activation with a small negative slope of  $\alpha=0.01$. In the last layer, separate convolutional layers are used for the two tasks, where $D_{src}$ returns a scalar score for computing the loss according to Wasserstein GAN~\cite{arjovsky_wasserstein_2017}, and $D_{attr}$  returns a vector of probabilities for each attribute class. 

{\bf Face Matcher.} Lastly, the auxiliary face matcher is adapted from the publicly available pre-trained VGG-Face CNN model that receives input face images of size $224\times 224\times 3$ and computes their face descriptors of size $2622$~\cite{parkhi_deep_2015}.

\subsection{Datasets}\label{sed:datasets}
\label{sec:datasets}

We have used five datasets in this study: CelebA~\cite{liu_deep_2015}, MORPH~\cite{ricanek_morph_2006}, MUCT~\cite{milborrow_muct_2010}, RaFD~\cite{langner_presentation_2010}, and UTK-face~\cite{zhifei_utkface_2017}. Table~\ref{tab:datasets-attrib-counts} shows the number of examples in each dataset, including the number of examples for each face attribute. Since the race label distribution in CelebA is heavily skewed towards Caucasians, and MORPH is heavily skewed towards persons with African ancestry, we combined CelebA and MORPH for training. Both the CelebA and MORPH datasets are split into training and evaluation sets in a subject-disjoint manner. The two training subsets from CelebA and MORPH are merged to train the PrivacyNet model with a relatively balanced race distribution. The other three datasets, MUCT, RaFD, and UTK-face are used only for evaluation. While all five datasets provide provide binary attribute gender labels~\footnote{In this paper we treat gender as a binary attribute with two labels, male and female; however, it must be noted that societal and personal interpretation of gender can result in many more classes.}, each dataset lacks the ground-truth labels for at least one of the other attributes, age or race. 

\begin{table*}
\begin{center}
\caption{Overview of datasets used in this study, with the number of face images corresponding to each attribute. Samples which belong to a race other than the two categories shown below, as well as those whose age-group could not be determined, are omitted.}
\label{tab:datasets-attrib-counts}
\begin{threeparttable}
%\begin{center}
%\scalebox{1.0}{ 
\centering
\small
\begin{tabular}{l | cc | cc | ccc}
 %\multicolumn{4}{c}{\bf Ensemble1: Regular} 
 \toprule
 %\multirow{2}{*}{ {\bf Dataset} } {\bf Dataset}  & {\bf \#male}& {\bf \#female}&   \multirow{2}{*}{ {\bf Usage} } \\
 \multirow{2}{*}{ {\bf Dataset} } & \multicolumn{2}{c|}{\bf Gender} & \multicolumn{2}{c|}{\bf Race}& \multicolumn{3}{c}{\bf Age groups}\\
 & {\bf Male}& {\bf Female}& {\bf African-descent} & {\bf Caucasian} & {\bf Young} & {\bf Midle-aged} & {\bf Old}\\
%& {images} & {images} & \\
\midrule %I remived \num[group-separator={,}]
CelebA & 84,434 & 118,165 & 11,119 & 142,225 & 79,848 & 91,373 & 16,337\\ 
MORPH &  47,057 & 8,551 & 42,897 & 10,736 & 25,009 & 26,614 & 3,985\\ 
MUCT & 1,844 & 1,910 & 1,030 & 1,480 & 1,326 & 1,807 & 620\\ 
RaFD & 1,008 & 600 & 0 & 1,608 & 1,276 & 332 & 0\\ 
UTK-face & 12,582 & 11,522 & 4,558 & 10,222 & 12,980 & 6,068 & 5,056\\
\bottomrule 
\end{tabular} 
\end{threeparttable}
\end{center}
\end{table*}

{\bf Gender Attribute:} All the five datasets considered in this study provide ground-truth labels for the gender attribute. Furthermore, since gender is a well-studied topic, there are several face-based gender predictors available for evaluation. In this study, we have considered three gender classifiers for evaluation: a commercial-off-the-shelf software G-COTS, IntraFace~\cite{de_la_torre_intraface_2015_long}, and AFFACT~\cite{gunther_affact_2017}.

{\bf Race Labels:} We consider binary labels for race: Caucasians and African descent. Samples that do not belong to these two race groups are omitted from our study since the other race groups are under-represented in our training datasets. We have used the ground-truth labels provided in the MORPH and UTK-face datasets, but for the other three datasets, we labeled the samples in multiple stages. First, an initial estimate of the race attribute is computed using commercial software R-COTS. Next, the predictions made by R-COTS from all samples of the same subject are aggregated, and subjects that show discrepant predictions for different samples are visualized and the discrepant labels are manually corrected. Finally, one random sample from every subject is visually inspected to verify the predicted label. Furthermore, note that since RaFD did not have any sample from the African-descent race group, we did not use this dataset for race prediction analysis. 

{\bf Age Information:} The ground-truth age information is only provided in the MORPH and UTK-face datasets. Therefore, for the remaining datasets (CelebA, MUCT, and RaFD) we used the commercial-off-the-shelf A-COTS software to obtain the class labels of the original images. For the evaluation of our proposed model, we use the Mean Absolute Error (MAE) metric to measure the change in predicted age on the output images of PrivacyNet from the predicted age on the original face images. Therefore, the combination of all five datasets shows both changes in age prediction with respect to the original (for CelebA, MUCT, and RaFD) as well as the ground-truth age values (for MORPH and UTK-face datasets). For training the PrivacyNet model, we create three age groups based on the age values:

\begin{equation}
    y_{\text{age}} = \left\{\begin{array}{cr}
         0 &  \text{age}\le 30;\\
         1 &  30 < \text{age} \le 45;\\
         2 & 45 < \text{age}. 
    \end{array}\right.
\end{equation}

Due to the non-stationary nature of patterns in face aging~\cite{chen_using_2017,niu_ordinal_2016}, creating age groups does not fully capture the non-linearity in the textural changes. However, this scheme is consistent with the treatment of the other two attributes, gender and age. Further, it should be emphasized that our objective is \emph{not} to synthesize face images in particular age groups (which is known as age synthesis); instead, the goal of the proposed method is to disturb the performance of arbitrary age predictors.

{\bf Identity Information:} For matching analysis, we exclude the UTK-face dataset since the subject information is not provided. We used three face matchers, a commercial-off-the-shelf software M-COTS, and two publicly available face matchers DR-GAN~\cite{tran_disentangled_2017} and SE-ResNet-50~\cite{hu_squeeze_2018}  (SE-Net for short) which were trained on the VGGFace2 dataset~\cite{cao_vggface2_2018}.

A summary of the datasets and the number of subjects and samples in each dataset is provided in Table~\ref{tab:dataset-info}.

\begin{table}[]
    \centering
    \caption{Summary of the datasets used in this study, with the number of subjects and samples in the train-test partitions. The ``Excluded Experiments" column indicate datasets that were removed from an experiment for the reasons given in the text.}
    \label{tab:dataset-info}
    \scalebox{0.8}{
    \begin{tabular}{c|cc|cc|c}
        \toprule
        \multirow{2}{*}{Datasets} & \multicolumn{2}{c|}{Train} & \multicolumn{2}{c|}{Test} & Excluded   \\
         & \# Subj & \# Samples & \# Subj & \# Samples & Experiments \\ \midrule
        CelebA & {8,604} & \num[group-separator={,}]{150530} &167 & {2,795}& -- \\ 
        MORPH  & \num[group-separator={,}]{11176} & \num[group-separator={,}]{45512} & {1,968}& {8,038}& -- \\
        MUCT   & -- & -- & 185&{2,508} & -- \\
        RaFD   & -- & -- & 67& {1,608}& Race \\
        UTK-face & -- & -- & NA& \num[group-separator={,}]{14182}& Matching \\
        \bottomrule
    \end{tabular}}
\end{table}

\section{Experimental Results}\label{sec:exp-results}

The proposed PrivacyNet model is trained on the joint training subsets of CelebA and MORPH as explained in Section~\ref{sec:datasets}. Due to the memory-intensive training process, we used a batch-size of $16$. The models were trained for $200,000$ iterations. The optimal hyperparameter settings for the weighting coefficients of the attribute loss terms were $\lambda_{attr,d}=1$ and $\lambda_{attr,g}=4$. The matching term coefficient was set to $\lambda_{m}=4$, and the hyperparameter for the  reconstruction term was set to $\lambda_{rec}=4$. After training the PrivacyNet model, both the discriminator and the auxiliary face matcher subnetworks are discarded and only the generator is used for transforming the unseen face images in the evaluation datasets.

\begin{figure*}
\begin{center}
   \includegraphics[width=0.85\linewidth]{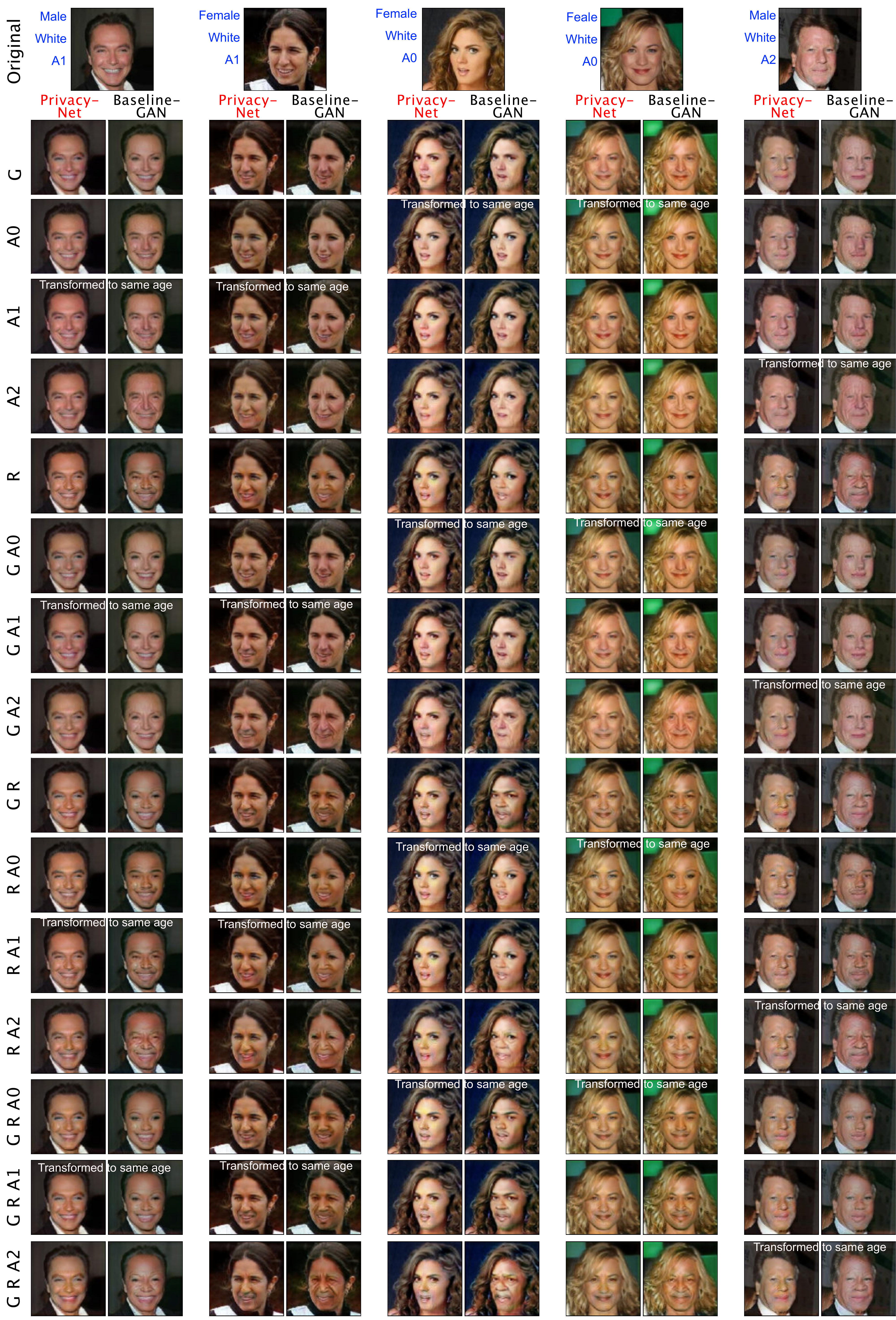}
\end{center}
   \caption{Five example face images from the  CelebA dataset along with their transformed versions using PrivacyNet and baseline-GAN models. The rows are marked by their selected attributes: G: gender, R: race, and A: age, where the specific target age group is specified as A0 (young), A1 (middle-aged), or A2 (old). }
\label{fig:example-outputs}
%\label{fig:onecol}
\end{figure*}

Additionally, we also trained a cycle-GAN model~\cite{choi_stargan_2017}, without the auxiliary face matcher, as a baseline to study the effects of the face matcher. The cycle-GAN model is trained using the same protocol that was described for training PrivacyNet. In the remainder of this paper, we will refer to this method as ``baseline-GAN''. The transformations of five different example images from the CelebA-test dataset are shown in Fig.~\ref{fig:example-outputs}.  

%\begin{landscape}
%\end{landscape}

The following subsections summarize the results of the experiments and analyze how the performance of the
attribute classifiers and face matchers is affected by the face attribute perturbations via PrivacyNet.
 
\subsection{Perturbing Facial Attributes}

The performance assessment of the proposed PrivacyNet model involves three objectives: 
\begin{enumerate}
    \item when an attribute is selected to be perturbed, the performance of unseen attribute classifiers must decrease;
    \item the attribute classifiers should retain their performance on attributes that are not selected for perturbation;
    \item in all cases, the performance of unseen face matchers must not be drastically affected.
\end{enumerate}

We conducted several experiments to assess whether the proposed PrivacyNet model meets these objectives.

\begin{figure*}[t]
\begin{center}
   \includegraphics[width=1.0\linewidth]{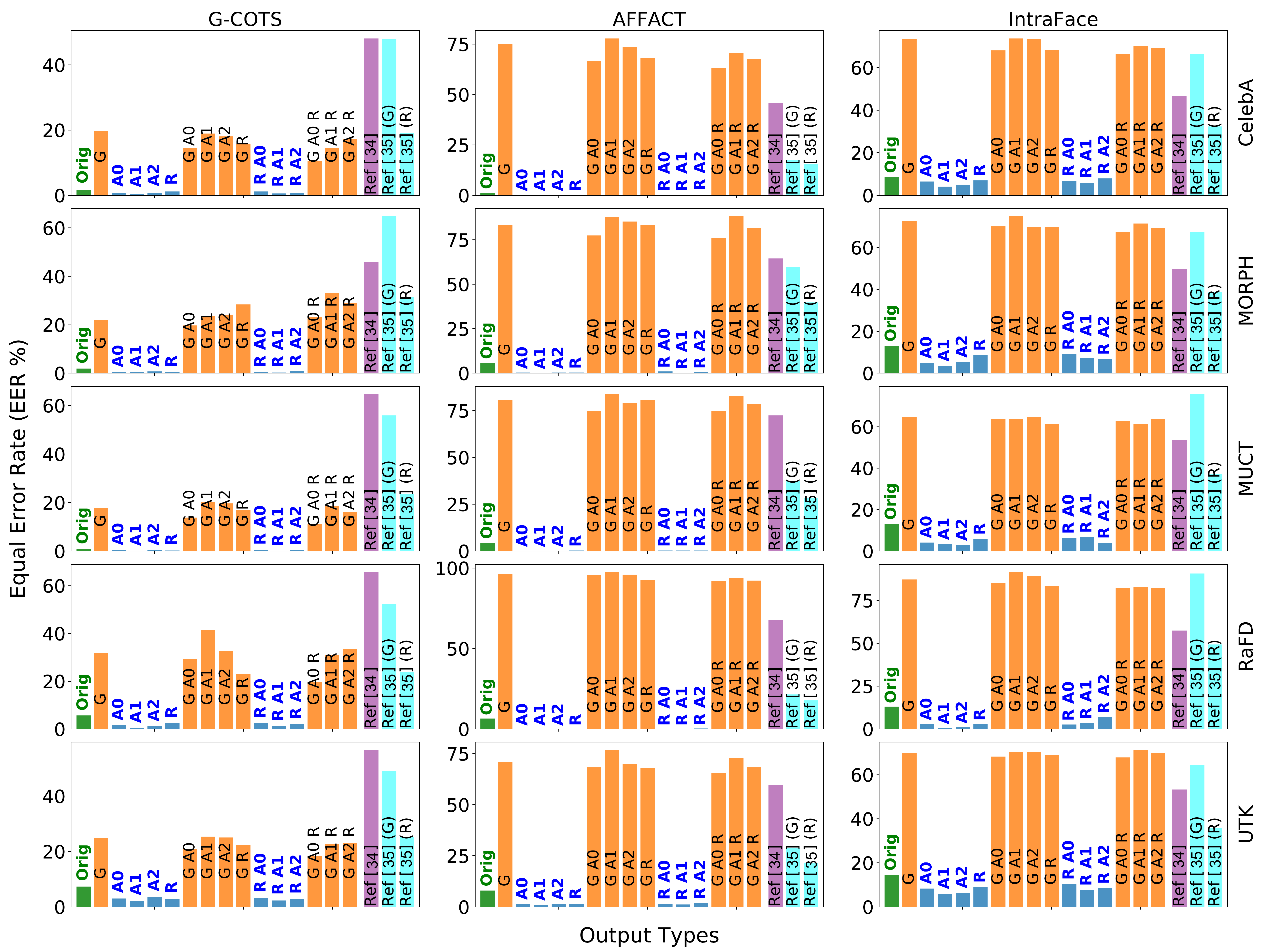}
\end{center}
   \caption{Performance of three gender classifiers -- G-COTS, AFFACT, and IntraFace -- on original images as well as different outputs of the proposed model (the larger the difference the better). The results of a face mixing approach, as described in~\cite{othman_privacy_2014}, are also shown. Different outputs are marked by their selected attributes: G: gender, R: race, and A: age, where the specific target age group is abbreviated as A0 (young), A1 (middle-aged), and A2 (old). The outputs of PrivacyNet, where the gender attribute is selected for perturbation, are shown in orange, and the rest are shown in blue.}
\label{fig:gender-eer-plots}
%\label{fig:onecol}
\end{figure*}

{\bf Gender Classification Performance:}
We considered three gender classifiers: a commercial-off-the-shelf software (G-COTS), AFFACT~\cite{gunther_affact_2017} and IntraFace~\cite{de_la_torre_intraface_2015_long}. For this comparison study, all five evaluation datasets listed in Table~\ref{tab:dataset-info} were considered. 
The performances
of the different gender classifiers on the original and perturbed images are measured using the Equal Error Rate (EER); the results are shown in
Fig.~\ref{fig:gender-eer-plots}. For a given image, PrivacyNet can produce up to 15 distinct outputs, depending on the
combination of attributes that are selected for perturbation.

The EER results shown in Fig.~\ref{fig:gender-eer-plots} indicate that PrivacyNet increases the error rate of the cases where the gender attribute is willfully perturbed, which is desired. At the same time, it can preserve the performance of gender classifiers when gender information is to be retained. The EER of gender classification using G-COTS software on gender-perturbed outputs increases to 20-40\%, and the EER of gender classification using AFFACT and IntraFace on these outputs surpasses 60\%. 
Comparisons between the gender prediction results on
the outputs of PrivacyNet and the outputs of the face-mixing approach by
Othman and Ross~\cite{othman_privacy_2014}, as well as the model by Sim and
Zhang~\cite{sim_controllable_2015}, show that in case of G-COTS, the PrivacyNet results are superior in terms of increasing the EER (Fig.~\ref{fig:gender-eer-plots}).

Note that we did not include the results of the GAN model in Fig.~\ref{fig:gender-eer-plots} for readability sake. However, we observed that the GAN model shows larger deviations (which is advantageous) in cases where gender was intended to be perturbed.
This is expected since the GAN model does not have the constraints from the auxiliary face matcher. Therefore, there is more flexibility for modifying the patterns of the face. However, a disadvantage of the GAN model is that it also significantly degrades the matching utility as shown in Section~\ref{sec:results-matching}.

\begin{figure*}[t]
\begin{center}
   \includegraphics[width=0.75\linewidth]{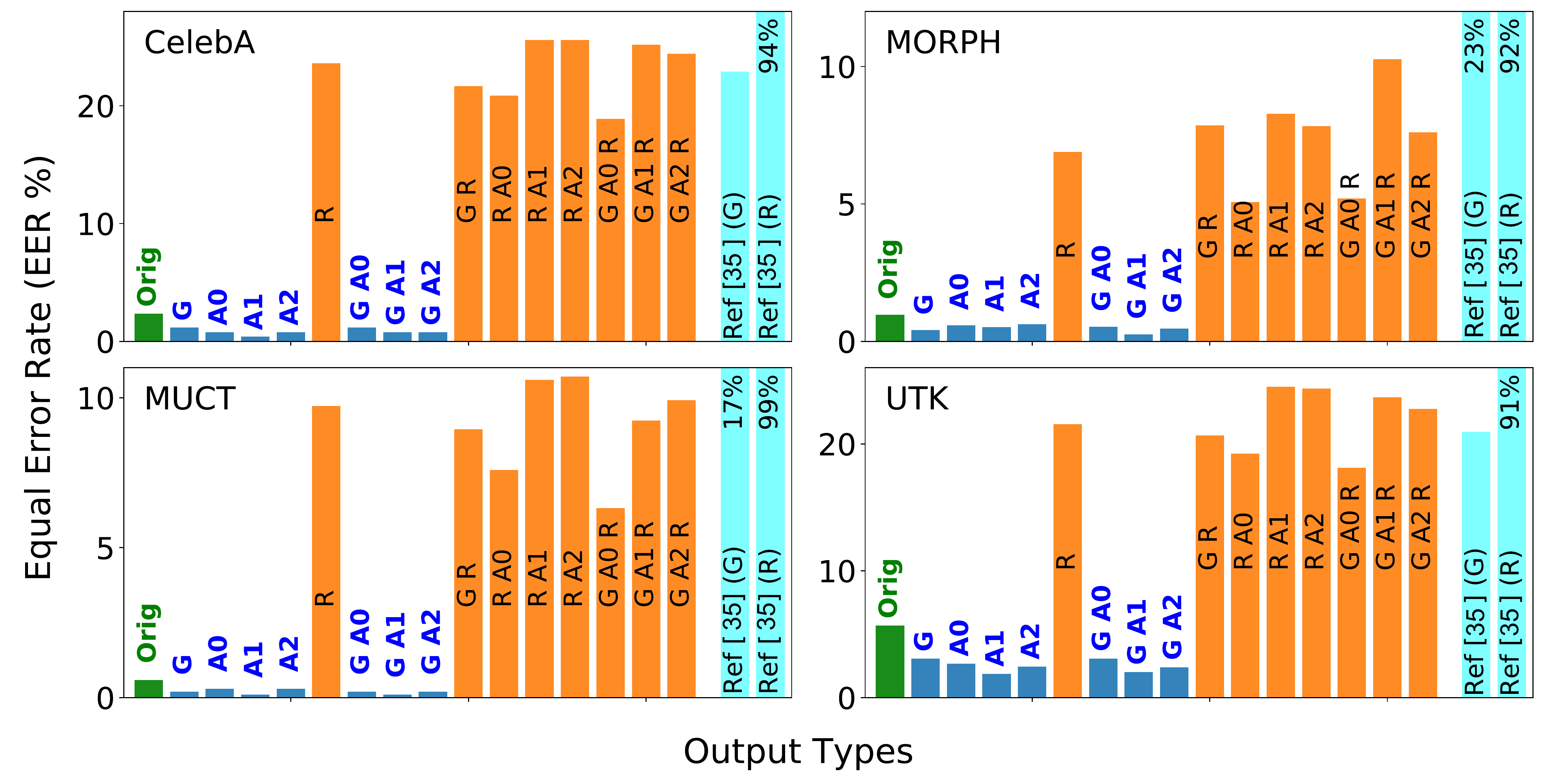}
\end{center}
   \caption{Performance of the race classifier, R-COTS, on original images as well as different outputs of the proposed model. Different outputs are marked by their selected attributes: G: gender, R: race, and A: age, where the specific target age group is denoted as A0, A1, and A2 (the larger the difference the better). The outputs of PrivacyNet, where the race attribute is selected for perturbation, are shown in orange, and the rest are shown in blue.}
\label{fig:race-eer-plots}
%\label{fig:onecol}
\end{figure*}

{\bf Race Prediction Performance:}
We conducted the race prediction analysis using a commercial-off-the-shelf software, R-COTS. 
Similar to the gender classification experiments, we show the EER of race classification on original images as well as the different outputs of the PrivacyNet model in Fig.~\ref{fig:race-eer-plots}. 
Since the face mixing approach proposed in~\cite{othman_privacy_2014} was only formulated for gender and not race perturbations, we did
not include it in this section.

The EER results in Fig.~\ref{fig:race-eer-plots} show that PrivacyNet successfully meets the objectives of our study for confounding race predictors. The outputs where race is not intended to be perturbed (shown in blue) exhibit low EER values similar to the EER obtained from the original images ($\text{EER} \sim 1\%$). On the other hand, when race is selected to be perturbed, the EER values increase significantly ($\text{EER} \sim 20\%$ for CelebA and UTK-face, and $\text{EER} \sim 10\%$ for MORPH and MUCT datasets). The results of separately perturbing gender and race using the controllable face privacy method proposed in~\cite{sim_controllable_2015} are also shown for comparison. When the race attribute is perturbed according to~\cite{sim_controllable_2015}, the performance is slightly higher than our model. However, the disadvantage of the controllable face privacy method~\cite{sim_controllable_2015} is that when it perturbs the gender attribute, it also affects the race predictions.

\begin{figure*}[t]
\begin{center}
   \includegraphics[width=0.9\linewidth]{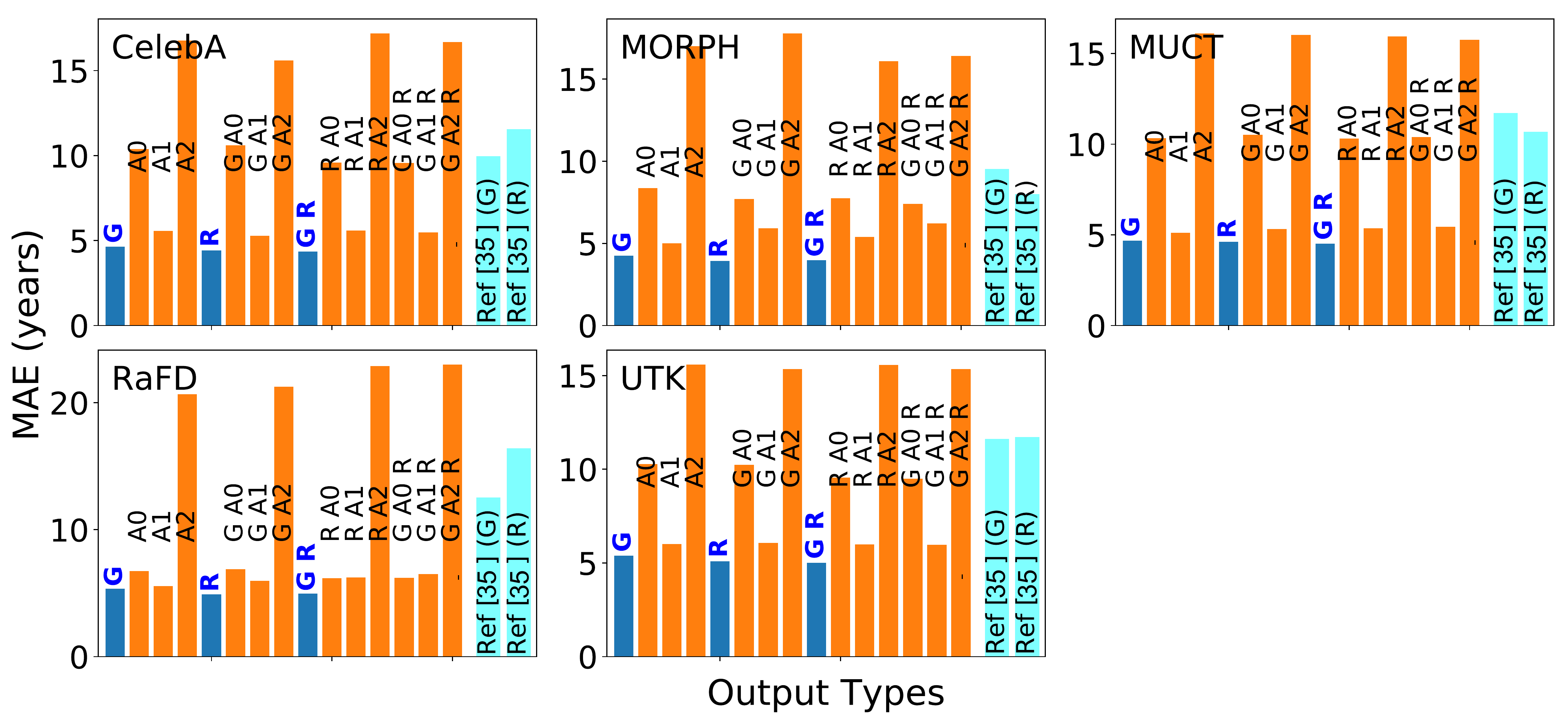}
\end{center}
   \caption{Change in age prediction of A-COTS on different outputs of the proposed model. This is with respect to the age predicted on original images for CelebA, MUCT and RaFD, and the ground-truth age values for MORPH and UTK-face. Different outputs are marked by their selected attributes: G: gender, R: race, and A: age, where the specific target age group is denoted as A0, A1 and A2. The outputs of PrivacyNet, where the age attribute is selected for perturbation, are shown in orange, and the rest are shown in blue.}
\label{fig:age-mae-plots}
\end{figure*}

{\bf Age Prediction Performance:}
To assess the ability of PrivacyNet for confounding age information, we used a commercial-off-the-shelf age predictor (A-COTS), which has shown remarkable performance across the different datasets tested in this study (Fig.~\ref{fig:age-mae-plots}).
%We have chosen a commercial-off-the-shelf age predictor software A-COTS which shows remarkable performance in age prediction across multiple datasets. 
We used the Mean Absolute Error (MAE) values in unit of years to measure the change in age prediction before and after perturbing the images (Fig.~\ref{fig:age-mae-plots}). As mentioned previously (Section~\ref{sec:datasets}), the ground-truth age values for three datasets -- CelebA, MUCT, and RaFD -- are not provided. Therefore, for these three datasets, the MAE values are computed as the difference between the age predictions on the output images and the predictions on the original images, while for the other two datasets, MORPH and UTK-face, the ground-truth values are used for computing the MAE values.

The results of age-prediction show that the MAE obtained from the outputs, where age is not meant to be perturbed, remains at approximately 5 years. However, when we  intend to modify the age of face images, using the label A2 results in the highest MAE (around 20 years for RaFD and 15 years for the other four datasets) compared to A0 and A1. 
A possible explanation for this observation is that, due the nature of the aging process, larger textural changes occur in face images belonging to A2.
%The primary reason for this behaviour is due to the nature of aging process where larger textural changes occur in face images for that age group (A2). 
The MAE of the A0 group  is
also relatively large (except for RaFD), which may be caused by the reversal of the textural changes.
%The MAE in A0 is also significantly large (except for RaFD) which is caused by reversing the textural changes. 
However, the results of the middle-age group (A1) is similar to the cases where we did not intend to modify the age.
We hypothesize that the small changes in A1 are also due to the non-stationary aspect of aging patterns; the age perturbations via the PrivacyNet model can potentially be improved by using an ordinal regression approach for age prediction.

%%%%%%%%%%%%%%%%%%%%%%%%%%%%%%%%%%%%%%%%%%%%%%%
\subsection{Retaining the Matching Utility of Face Images}
\label{sec:results-matching}

\begin{figure*}
\begin{center}
   \includegraphics[width=1.0\linewidth]{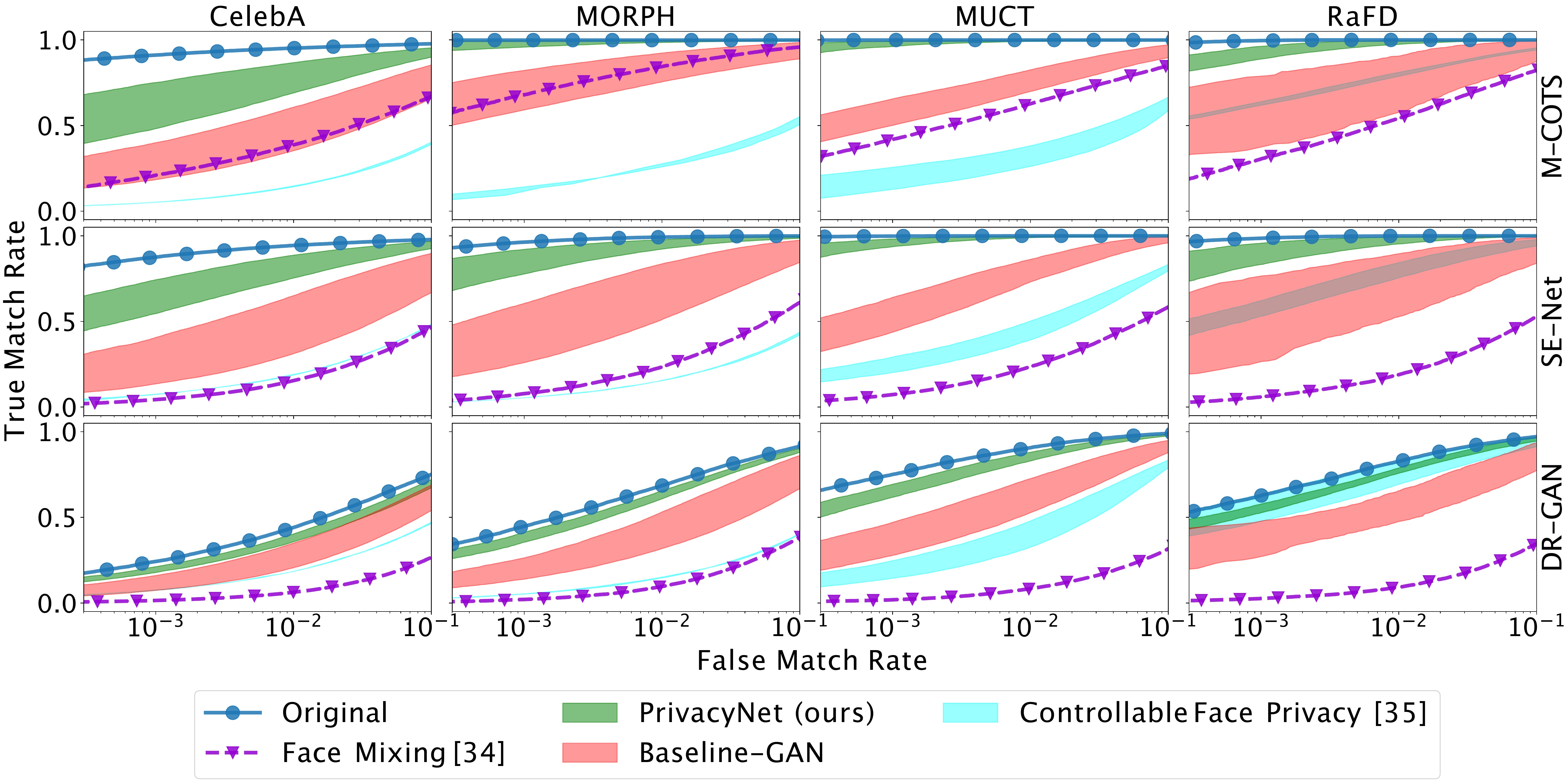}
\end{center}
   \caption{ROC  curves  showing  the  performance of unseen face matchers on the
original  images  for  PrivacyNet, the baseline-GAN model, face mixing~\cite{othman_privacy_2014} approach and the controllable face privacy~\cite{sim_controllable_2015} method. The results show that ROC curves of PrivacyNet have the smallest deviation from the ROC curve of original images suggesting that the performance of face matching is minimally impacted, which is desired.}
\label{fig:matching-roc-curves}
%\label{fig:onecol}
\end{figure*}

\begin{figure*}
\begin{center}
   \includegraphics[width=1.0\linewidth]{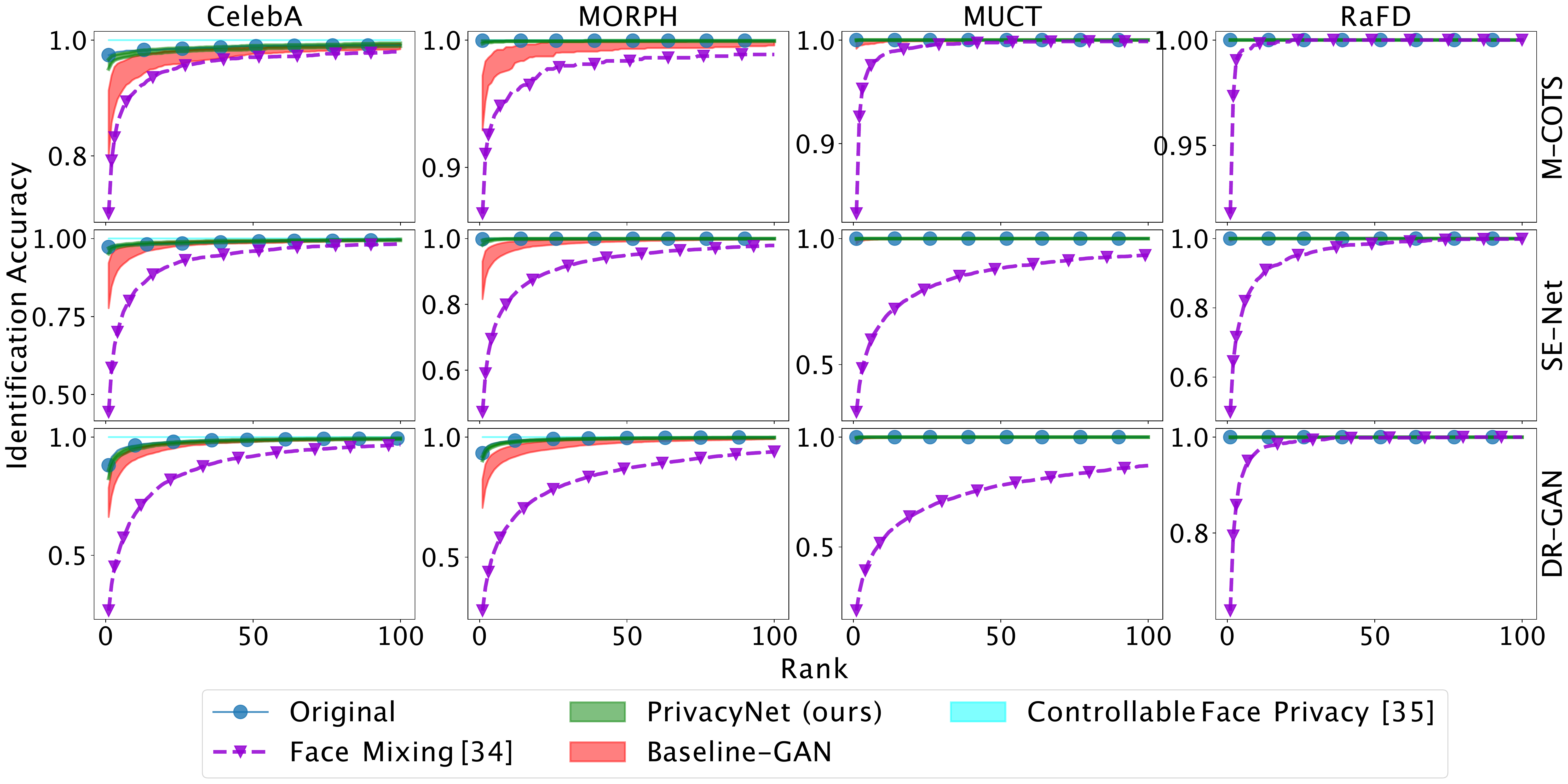}
\end{center}
   \caption{CMC  curves  showing  the identification accuracy of unseen face matchers on the original images. Also shown is the range of CMC curves for the PrivacyNet model and the baseline-GAN model, along with that of the face mixing~\cite{othman_privacy_2014} and controllable face privacy~\cite{sim_controllable_2015} approaches. It must be noted that in cases where the results of PrivacyNet or GAN are not visible, the curves overlapped with the CMC curve of the original images: this means that there was no change in matching performance at all (which is the optimal case). The results confirm that transformations made by PrivacyNet preserve the matching utility of face images.}
\label{fig:matching-cmc-curves}
%\label{fig:onecol}
\end{figure*}

Besides obfuscating soft-biometric attributes in face images, another objective of this work is to retain the recognition utility of all outputs of PrivacyNet. For this purpose, we conducted matching experiments using three unseen face matchers: 
commercial-off-the-shelf software (M-COTS) 
and two publicly available matchers,
%a commercial-off-the-shelf M-COTS, and two publicly available matchers, 
SE-ResNet-50 trained on the VGGFace2 dataset~\cite{cao_vggface2_2018} (SE-Net for short), and DR-GAN~\cite{tran_disentangled_2017}. Fig.~\ref{fig:matching-roc-curves} shows the ROC curves obtained from these matching experiments for four datasets -- CelebA, MORPH, MUCT, and RaFD. The UTK-face dataset is removed from this analysis since it does not contain subject information. Since PrivacyNet generated 15 outputs for each input face image, the minimum and maximum True Match Rate (TMR) values at each False Match Rate (FMR) value are computed and only the range of values for these 15 outputs are shown. Note that it is expected for the matching utility to be retained in all these 15 outputs. Similarly, the range of TMR values at each FMR obtained from the 15 different outputs of the GAN model that did not have the auxiliary face matcher for training, is also shown for comparison. The ROC curves of PrivacyNet are very close to the ones obtained from the original images for each dataset, compared to the baseline results, which both show significantly larger deviations. It is worth noting that the baseline-GAN is equivalent to removing the matching loss term $\mathcal{L}_{G,m}$ from PrivacyNet. As shown in Figures~\ref{fig:example-outputs},\ref{fig:matching-roc-curves} and \ref{fig:matching-cmc-curves}-(``Baseline-GAN''), the PrivacyNet model produces more realistic-looking faces images without the matching loss term. However, removing the matching loss term results in a severe decline in matching performance, affecting both the true matching rate and identification accuracy (Figs.~\ref{fig:matching-roc-curves} and \ref{fig:matching-cmc-curves}). The coefficient $\lambda$ can be further tuned to control the trade-off between the performance of face-matching and obfuscating the soft-biometric attributes.

In addition to the ROC curves, we have also plotted the Cumulative Match Characteristics (CMC)~\cite{decann_relating_2013}, as shown in Fig.~\ref{fig:matching-cmc-curves}. According to the CMC curves, the results of PrivacyNet match very closely with the CMC curves obtained from the original images in all cases, which shows that PrivacyNet retains matching utility. 

It is worth noting that Ref.~\cite{sim_controllable_2015} has more favorable CMC curves than the other methods evaluated in this study. A plausible explanation is that Ref.~\cite{sim_controllable_2015} aligns and normalizes its inputs to a reference face image, which significantly reduces the intra-class variations. This reduction of intra-class variations increases the number of true positives. However, it also increases the number of false positives, thereby deteriorating the {\it ROC} performance. One may argue that the difference in performance could be due to the different training datasets that were used to train our model and that of Ref.~\cite{sim_controllable_2015}, and, perhaps, re-training Ref.~\cite{sim_controllable_2015} would be necessary for a fair comparison. However, we note that we used the original model for Ref.~\cite{sim_controllable_2015}, which was constructed from a carefully curated dataset, and the original authors of~\cite{sim_controllable_2015} recommended against retraining.

\section{Conclusion}
%In this work, we designed a special neural network model coined PrivacyNet for imparting multi-attribute privacy
In this work we designed PrivacyNet, which is a deep neural network model for imparting multi-attribute privacy to face images including age, gender, and race attributes. PrivacyNet utilizes a Semi-Adversarial Network (SAN) module combined with Generative Adversarial Networks (GANs) to perturb an input face image, where certain attributes are obfuscated selectively, while other face attributes are preserved. Most importantly, the matching utility of face images from these transformations is preserved. 
Experimental results using three unseen face matchers as well as three unseen attribute classifiers show the efficacy of our proposed model in perturbing such attributes, while the matching utility of face images is not adversely impacted.

Although generating visually realistic images was not the primary objective of this work, we note that the modified images from the proposed model may have some artifacts. As a result, a human observer might be able to distinguish between perturbed face images and non-modified ones. We intend to study the effect of perturbations on human observers in future work and plan to design solutions capable of creating more realistic-looking face images while satisfying the objective of this work, viz., maintaining good face matching performance.

% or


%\appendices

%\section{}

\ifCLASSOPTIONcompsoc
  \section*{Acknowledgments}
\else
  \section*{Acknowledgment}
\fi

We would like to thank Pranavan Theivendiram and Terence Sim for kindly providing a Python API for controllable face privacy~\cite{sim_controllable_2015}, which was used for the comparison studies.
In addition, the authors would like to thank the US National Science Foundation (Grant Number $1618518$), as well as computational resources that were acquired through funds from the University of Wisconsin Alumni Research Foundation.

\ifCLASSOPTIONcaptionsoff
  \newpage
\fi



%\bibliographystyle{IEEEtran}
{\small
\bibliographystyle{IEEEtran}
\balance
\bibliography{egbib}
}

\end{document}